\documentclass{IEEEtran}

\usepackage{glossaries}

\newacronym{mcts}{MCTS}{Monte Carlo Tree Search}
\newacronym{mcs}{MCS}{Monte Carlo Search}
\newacronym{ismcts}{IS-MCTS}{Information Set-\acrlong{mcts}}
\newacronym{uct}{UCT}{Upper Confidence bound for Trees}
\newacronym{ucb}{UCB}{Upper Confidence Bound}
\newacronym{ga}{GA}{Genetic Algorithm}
\newacronym{tom}{ToM}{Theory of Mind}
\usepackage{tabularx}
\usepackage{graphicx}
\usepackage{cleveref}
\usepackage{verbatim}
\usepackage{nameref} 

\usepackage{subcaption}
\usepackage{fancyhdr}

\usepackage{todonotes}
\usepackage{fancyhdr}

\usepackage{tikz}
\usetikzlibrary{arrows}

\tikzset{
	treenode/.style = {align=center, inner sep=0pt, text centered,
		font=\sffamily},
	arn_n/.style = {treenode, circle, white, font=\sffamily\bfseries, draw=black,
		fill=black, text width=1.5em},
	arn_r/.style = {treenode, circle, red, draw=red, 
		text width=1.5em, very thick},
	arn_x/.style = {treenode, rectangle, draw=black,
		minimum width=0.5em, minimum height=0.5em}
}

\graphicspath{{}{results/}}

\fontsize{10}{12}

\title{Evaluating and Modelling {\em Hanabi}-Playing Agents}
\author{
	\IEEEauthorblockN{	
		Joseph Walton-Rivers\IEEEauthorrefmark{1} \and Piers R. Williams\IEEEauthorrefmark{2} \and Richard Bartle\IEEEauthorrefmark{3} \and Diego Perez-Liebana\IEEEauthorrefmark{4} \and Simon M. Lucas\IEEEauthorrefmark{5} 
	}
	\IEEEauthorblockA{
		\small{
			\\School of Computer Science and Electronic Engineering,\\University of Essex, Colchester \\ CO4 3SQ, UK\\
            Email: \{jwalto\IEEEauthorrefmark{1}, pwillic\IEEEauthorrefmark{2}, rabartle\IEEEauthorrefmark{3}, dperez\IEEEauthorrefmark{4}, sml\IEEEauthorrefmark{5}\}@essex.ac.uk
		}
	}
}

\begin{document}
\maketitle

          \thispagestyle{plain}
          \fancypagestyle{plain}{
            \fancyhf{} 
            \fancyfoot[L]{978-1-5090-4601-0/17/\$31.00~\copyright2017~IEEE} 
            \renewcommand{\headrulewidth}{0pt}
            \renewcommand{\footrulewidth}{0pt}
          }

\begin{abstract}
Agent modelling involves considering how other agents will behave, in order to influence your own actions. In this paper, we explore the use of agent modelling in the hidden-information, collaborative card game {\em Hanabi}. We implement a number of rule-based agents, both from the literature and of our own devising, in addition to an \gls{ismcts} agent. We observe poor results from \gls{ismcts}, so construct a new, predictor version that uses a model of the agents with which it is paired. We observe a significant improvement in game-playing strength from this agent in comparison to \gls{ismcts}, resulting from its consideration of what the other agents in a game would do. In addition, we create a flawed rule-based agent to highlight the predictor's capabilities with such an agent.
\end{abstract}

\section{Introduction}
{\em Hanabi} is a co-operative, partially-observable~\cite{williamscooperative} board game which in 2013 won the prestigious {\em Spiel des Jahres} award for best board game of the year. For the reasons outlined below, it has featured in a number of recent academic publications. This paper explores whether the use of agent modelling can lead to an improvement in strength for agents playing the game.

{\em Hanabi} has a number of interesting features that make it a good choice for research in the field of agent modelling. Firstly, the domain is a co-operative one, in that the agents must work together to achieve a shared goal. This disfavours agents that behave greedily: for example, helping another player score a point is better than playing a risky card that might end the game. Secondly, its rules build in well-defined communication actions. These use a resource that regulates communication and must be managed by the agents. Finally, the game has hidden information, with no one player able to see the entire game state. This is a source of complexity for agents, because imperfect information needs to be reasoned about intelligently. Note that {\em Hanabi} has been proven to be NP-Complete even when players have perfect information~\cite{baffier2016hanabi}.

A number of rule-based approaches for designing agents that can play {\em Hanabi} have been presented in the literature, however there have been few attempts to employ more general strategies. In this paper, we go some way towards redressing this imbalance. Furthermore, because the use of information about other players' strategies can help to inform human players in co-operative games, we also explore whether or not such information could help guide our general agents' decisions.

\Cref{sec:intro:hanabi} describes the rules of the game of {\em Hanabi}.

\Cref{sec:ai} describes the agents that were implemented to play {\em Hanabi} under these rules.

\Cref{sec:method} describes how the agents were tested and evaluated.

\Cref{sec:results} presents the results of the tests.

\Cref{sec:discussion} discusses and explains our findings in the results.

\Cref{sec:future} discusses potential future AI-related work involving {\em Hanabi}.

\subsection{Hanabi}
\label{sec:intro:hanabi}
{\em Hanabi} is a co-operative game in which a team of two to five players attempts to complete five stacks of sequentially-numbered cards (one for each of the game's five suits).

The game is played with a deck of 50 cards, each possessing a suit and a rank. The suits are coloured white, yellow, green, blue and red. Within each suit, there are three cards of rank 1, two cards each of ranks 2, 3 and 4, and one card of rank 5. The game additionally features two types of token: an {\em information token} and a {\em life token}. The players collectively start the game with 3 life tokens and 8 information tokens. 

Every player begins with a randomly-dealt hand of five cards. Cards are held facing away, such that players can't see the suit or rank of their own cards but can see the suit and rank of the cards held by the other players. The cards not dealt out at the start are placed face down as a {\em draw deck}, which will be accessed during play.

Play proceeds with each player taking it in turn to perform an action of their choice. There are three different types of action available:

\vspace{0.4cm}

\begin{center}
\begin{tabularx}{\linewidth}{>{\bfseries}l X}
Tell & Select a player and point to all their cards of a given number or suit. This costs one information token. \\
Play & Choose a card from the player's own hand and play it.\\
Discard & Choose a card from the player's own hand and add it to the discard pile\\
\end{tabularx}
\end{center}
\vspace{0.3cm}

In a single Tell action, a set of cards can only be identified either by their suit or by their rank --- not by both. Furthermore, cards must be present in the hand to be identified: it is not permitted to state that another player has no cards of a given suit or rank.

Playing a card means adding it to the stack with matching suit. It is not required that the player know which stack it belongs to - for example, at the beginning of the game it is acceptable to blindly play a 1 that has been indicated to you. Each card in the stack must be of the correct suit and have a rank one greater than the card below (except for 1 cards, which are used to start a stack). If a card is played out of sequence, the group loses one life token. Completing a stack of cards associated with a given suit grants an additional information token (if the team does not already have the maximum number, eight).

Discarding is only permitted if there is at least one information token to be gained. This means that either a Tell action or a Discard action is always possible.

After either discarding or playing a card, the player draws a replacement card from the draw deck. Discarded cards are visible to all players. Discarding a card increments the number of information tokens up to the maximum. Once all cards in the draw deck have been drawn, all players get one more turn and then the game is considered to be over. The game also ends if the team uses up all of the life tokens.

Scoring is achieved by summing the top card of each stack that has been correctly played. The maximum possible score for the standard game is therefore 25, obtained by completing the stacks for all five suits. Remaining life or information tokens are not counted towards score in the standard game.

\subsection{Multi-agent domains}


Multi-agent domains can be categorised as either {\em centralised} or {\em distributed}. A centralised system features a single controller controlling multiple agents; a distributed system has each agent in the world  controlled by a separate controller. In this paper, we consider only the distributed approach.



Existing work in this space includes: attempting to reason about what the other agent knows using answer set programming~\cite{Baral:2010:UAS:1838206.1838243}; iterating on a plan that is communicated between agents~\cite{torreno2013fmap}; and attempting to use plan recognition to allow one agent to assist another in a planning task~\cite{dmap2016geib}.

Another possibility involves co-operative, multi-agent learning. Within this area, there have been attempts to learn models of teammates in order to make more informed decisions about which action to take. For a review of the literature, see Panait \& Luke~\cite{Panait2005}.

The use of embedding agent models into \gls{mcts} has previously been looked at by Barrett {\em et al} in the pursuit domain~\cite{barrett2011empirical}. They made the assumption that all agents except for their modelling agent would be using the same, fixed strategy, and embedded perfect knowledge of this strategies into their agent. One of their findings was that the system did not perform well with models that didn't represent the behaviour of the agent. 

The use of \gls{tom} (reasoning about what the other agents know and will do in a given situation) has proven useful in competitive games such as Rock Paper Scissors~\cite{de2013much}. In these games, higher-order \gls{tom} agents were able to out-perform lower-order \gls{tom} players.

\subsection{Co-ordination in Hanabi}

In {\em Hanabi} all agents have access to different information; because of this, a centralised approach to multi-agent planning would not make sense in this domain as private information must not be shared between agents.

The fact that the Tell action has an associated cost (an information token) means that information about a player's hand needs to be communicated efficiently. Also, because {\em Hanabi} players are limited to a set of well-defined communication actions, communication between them is very limited. This makes using communication between agents to co-ordinate their actions a challenging prospect --- which is one reason why {\em Hanabi} is increasingly becoming the object of research.

The understanding of other players' strategies forms a core component of a great number of games and has been studied widely~\cite{de2013much}. Existing {\em Hanabi} research assumes that all agents are playing the same pre-agreed strategy. The ability to reason about the actions that a player would take and their reasons for taking these actions can be used as part of the reasoning process of an agent. 

Our approach is to assume that we have access to a model which, given a state, will be able to return a possible action that an agent would perform in that state; if the agent may make multiple moves, then a single action from the set of possible actions will be returned. Given this model, we are able to incorporate the behaviour of the other agent into our model without understanding of that agent's reasoning process.

A point to note is that Tell actions can convey more information than just the obvious: because {\em all} cards of a given suit or rank must be identified, cards which are not identified therefore must not satisfy the criterion. This {\em negative information} can be used to inform the possible values for a given card. Negative information can add up over a few turns, providing enough information to determine what a card is --- or at least that it is playable. In the end game, such knowledge becomes very powerful.

Human players of {\em Hanabi} often make additional use of Tell actions. In particular, they can restrict their Tell actions by convention only to identify certain cards as playable. For example, suppose that Player 2 had the hand \{(R, 1), (B, 1), \ldots \} and the current stacks on the table were \{(R, 1), (B, 0), (G, 0), (W, 0), (Y, 0)\}. Player 1 may elect to tell Player 2 about the suit rather than the number, to avoid identifying the non-playable red card. Player 2 could then infer that the card being identified was indeed a playable card, as they would know that Player 1 would not have identified a non-playable card. As they were told the suit rather than the number, they could further infer that they have a non-playable 1 in their hand (although they would not know the location of this card). The use of information in this way requires an understanding of how the player will use the provided information as part of their policy.
 
\subsection{\acrlong{mcts}}
\label{sec:intro:mcts}
\gls{mcts}~\cite{browne2012survey} is a widely-used tree-search algorithm that can operate without domain-specific knowledge. This gives \gls{mcts} the {\em anytime} property: the algorithm can be stopped at any time and can provide an answer for the next move. Given more time, it will typically produce a more accurate answer. 

\begin{figure*}[!h]
\begin{center}
	\includegraphics[scale=0.75]{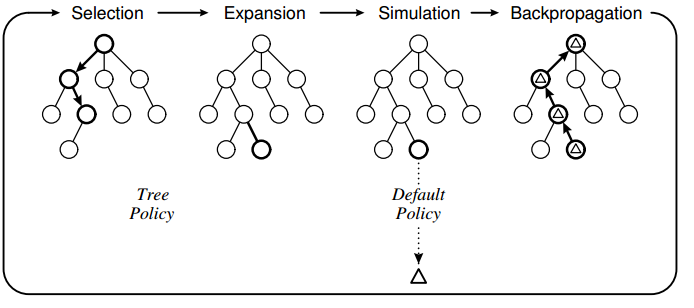}
\end{center}
\caption{The four steps of \acrlong{mcts}~\cite{chaslot2008progressive}}
\label{fig:mcts}
\end{figure*}

\gls{mcts} proceeds using multiple iterations of the four main stages shown in~\Cref{fig:mcts}. The iterations typically continue until a predetermined end condition is met, such as running out of time. In the selection stage, the current tree is traversed using the tree policy to select the best child of each node. In the expansion stage, a new node is added to the tree. In the main, simulation phase, a simulation ({\em rollout}) of future moves is undertaken from the state represented by the new node until an end condition is met. Moves are selected according to the default policy (which is often to select at random from all possible moves). In the backpropagation phase, nodes in the tree that were selected are updated with the result of the rollout.

\subsection{\acrlong{mcts} and \acrlong{tom}}
Zero-order theory of mind~\cite{de2013much} agents are capable of using an agent's history in order to inform future actions. A first-order theory of mind agent is capable of using a model of a zero-order theory of mind agent to inform its own future decisions. Our selection of \gls{mcts} for use in this domain came from a particular desire to find an algorithm that could be easily modified to operate with predictions of what other agents would do. This makes it a zero-order agent. 

This approach has been tested before in the Tiny Co-op domain~\cite{williams2015}  by Walton-Rivers, who found that prediction worked best with a deterministic agent that did as it was instructed~\cite{waltoncontrolling}. The Tiny Co-op domain is a simple, grid-based world containing a number of agents, goals, doors and buttons. Each agent must visit each goal individually for successful completion. Doors separate different areas in which the agents can move, and each door will only open if an agent is standing on its associated button. This forces the agents to co-operate to succeed overall.

While \gls{mcts} was a good performer in Tiny Co-op when paired with itself (and even with random agents), it struggled when trying to co-operate with a particular agent that was designed to follow direction indications. Essentially, this {\em follower} agent moved to where it was instructed to move, but \gls{mcts} didn't pick up on this. The root cause was that it didn't model such behaviour in its search tree, leading to inaccurate states in the majority of the search space. The author added agent modelling to \gls{mcts} and found that the performance of \gls{mcts} when paired with the follower agent improved significantly. In this paper we used this approach to create a {\em Hanabi}-playing agent to assess the effectiveness of agent modelling in this domain.

\subsection{Previous research}
\label{sec:intro:prev}

\subsubsection{Imperfect Information AI}
\label{sec:intro:prev:imperfect}
Games with imperfect information are a complex challenge for AI. Poker is often chosen as an application, because it is a game that many people are familiar with on some level. Poker contains an unusual dynamic for games, as a strong player doesn't so much play the game as play the opponents. Winning requires a player to understand their opponents and to adopt a strategy that will counter their strengths while exploiting their weaknesses. Rule-based agents feature strongly in this, as do simulation-based agents such as \gls{mcts}. Poker has been extensively studied --- see the review conducted by Rubin \& Watson~\cite{rubin2011computer}; one of their notable finds was that a simulation-based approach is inferior to the formula-based approach, despite expectations.

Whitehouse {\em et al}~\cite{whitehouse2011determinization} looked into using \gls{mcts} for the card game Dou Di Zhu, which (like {\em Hanabi}) also features imperfect information. Here, they apply determinisation and \gls{ismcts} to the problem and conclude while the \gls{ismcts} is superior in some cases, no overall difference was observed.

\subsubsection{Hanabi AI}
\label{sec:intro:prev:hanabi}
There has been a small amount of research into using artificial intelligence techniques to play {\em Hanabi}. Osawa~\cite{Osawa2015} devised a number of rule-based agents for the 2-player version of the game, the mechanisms for which are described in \Cref{sec:agent-internal,sec:agent-outer}. Osawa found that the incorporation of consideration of the other agent's  strategy and why they did what they did allows an agent to perform better than do the other non-cheating agents.

Cox~\cite{cox2015make} derives strategies for the game of {\em Hanabi} using the \textit{hat guessing game} as inspiration. The agents all use an agreed encoding strategy to indicate what any particular Tell action specifically means, enabling them to co-operate so as to work around the limited view of their own hands. The encoding strategy does require the 5 player version of the game, however, as it won't work unless the hand size matches the number of other players in the game. We considered using this agent in the tests, as its unique strategy could have been the perfect test for agent modelling. However, there is an issue with the encoding strategy: every agent must know what every other agent has in their hands. This is cannot be used in agent modelling. If the \nameref{sec:agent-predictor} is agent 1, then it has access to the hands of agents 2, 3, 4 and 5 --- which in {\em Hanabi} it does indeed have. Unfortunately, its internal copy of agent 2 needs access to the hands of agents 1, 3, 4 and 5. Agent 1 cannot give this information without breaking the rules of {\em Hanabi}. For this reason, we did not run tests with this agent.

Van den Bergh {\em et al}~\cite{mi2015hanabi} analyse {\em Hanabi} and define a number of rules for the game. The amount of time it would take to test every possible combination of these rules was too large, however, so they used an iterative approach to explore the search space intelligently. They note that some rules are far more effective than others, as well as observing that a risk-taking rule does have some value. They found that the use of a Discard action when there is a possible hint is not optimal. In a follow-up paper~\cite{vanaspects}, the authors present their best rule-based agent along with one using a Monte Carlo search.

\section{AI}
\label{sec:ai}
A number of the controllers used in this experiment were implemented as production rule agents. Many of these share individual rules, so each rule will be described here independently. All rules have additional pre-conditions that ensure they can only fire if it is legal to do so within the game rules (for example, a Discard action would necessitate a check that an information token was available). To avoid verbosity, we assume that the rules of {\em Hanabi} will be properly followed (so, for example, if the rule says to inform a player about a card, then the player will also be informed about other cards that satisfy the Tell's stated criterion).

\begin{itemize}
\item{\textbf{PlaySafeCard: }Plays a card only if it is guaranteed that it is playable}
\item{\textbf{OsawaDiscard: }Discards a card if it cannot be played at the end of the turn. This will discard cards that we know enough about to disqualify them from being playable. For example, a card with an unknown suit but a rank of 1 will not be playable if all the stacks have been started. This rule also considers cards that can not be played because their pre-requisite cards have already been discarded.}
\item{\textbf{TellPlayableCard: }Tells the next player a random fact about any playable card in their hand.}
\item{\textbf{TellRandomly: }Tells the next player a random fact about any card in their hand.}
\item{\textbf{DiscardRandomly: }Randomly discards a card from the hand.}
\item{\textbf{TellPlayableCardOuter: }Tells the next player an unknown (to that player) fact about any playable card in their hand.}
\item{\textbf{TellUnknown: }Tells the next player an unknown fact about any card in their hand.}
\item{\textbf{PlayIfCertain: }Plays a card if we are certain about which card it is and that it is playable.}
\item{\textbf{DiscardOldestFirst: }Discards the card that has been held in the hand the longest amount of time.}
\item{\textbf{IfRule($\lambda$) Then (Rule) Else (Rule): } Takes a Boolean $\lambda$ expression and either one or two rules. The first rule will be used if the $\lambda$ evaluates to true. If it is false, and a second rule was provided, then that will be used instead.}
\item{\textbf{PlayProbablySafeCard($Threshold \in[0, 1]$): } Plays the card that is the most likely to be playable if it is at least as probable as {\em Threshold}.}
\item{\textbf{DiscardProbablyUselessCard($Threshold \in[0, 1]$): } Discards the card that is most likely to be useless if it is at least as probable as {\em Threshold}.}
\item{\textbf{TellMostInformation($New? \in[True, False]$): } Tells whatever reveals the most information, whether this is the most information in total or the most new information.}
\item{\textbf{TellDispensable: } Tells the next player with an unknown dispensible card the information needed to correctly identify that the card is dispensible. This rule will only target cards that can be identified to the holder as dispensible with the addition of a single piece of information.}
\item{\textbf{TellAnyoneAboutUsefulCard: } Tells the next player with a useful card either the remaining unknown suit of the card or the rank of the card.}
\item{\textbf{TellAnyoneAboutUselessCard: } Tells the next player with a useless card either the remaining unknown suit of the card or the rank of the card.}
\end{itemize}

\subsection{Agents}
\label{sec:agents}
\subsubsection{Legal Random} 
\label{sec:agent-legal_random}
This agent makes a move at random from the set of legal actions available to it at any given time step.

\subsubsection{Internal}
\label{sec:agent-internal}
This is a clone of the agent presented by Osawa that shares the same name. It features memory of the information it has been told about its own hand but does not remember information about what other players have been told. The rules used in order are:

\begin{itemize}
	\item{PlaySafeCard}
	\item{OsawaDiscard}
	\item{TellPlayableCard}
	\item{TellRandomly}
	\item{DiscardRandomly}
\end{itemize}

\subsubsection{Outer}
\label{sec:agent-outer}
This is a clone of the agent presented by Osawa with the same name. It features knowledge of what the other agents have been told already, to avoid repeating Tell actions. The rules used in order are:

\begin{itemize}
	\item{PlaySafeCard}
	\item{OsawaDiscard}
	\item{TellPlayableCardOuter}
	\item{TellUnknown}
	\item{DiscardRandomly}
\end{itemize}

\subsubsection{Cautious}
\label{sec:agent-cautious}
This is an agent derived from human gameplay. The agent plays cautiously, never losing a life. The rules used in order are:

\begin{itemize}
	\item{PlayIfCertain}
	\item{PlaySafeCard}
	\item{TellAnyoneAboutUsefulCard}
	\item{OsawaDiscard}
	\item{DiscardRandomly}
\end{itemize}

\subsubsection{IGGI}
\label{sec:agent-iggi}
This agent is a modification of Cautious. The alteration to a deterministic Discard function greatly aids the predictability of this player. The rules used in order are:

\begin{itemize}
	\item{PlayIfCertain}
	\item{PlaySafeCard}
	\item{TellAnyoneAboutUsefulCard}
	\item{OsawaDiscard}
	\item{DiscardOldestFirst}
\end{itemize}

\subsubsection{Piers}
\label{sec:agent-piers}
This is an agent designed to use IfRules to improve the overall score. Otherwise, it is similar to \nameref{sec:agent-iggi}. The rules used in order are:

\begin{itemize}
	\item{IfRule (lives $>$ 1 $\land$  $\lnot$deck.hasCardsLeft) Then (PlayProbablySafeCard(0.0))}
	\item{PlaySafeCard}
	\item{IfRule (lives $>$ 1) Then (PlayProbablySafeCard(0.6))}
	\item{TellAnyoneAboutUsefulCard}
	\item{IfRule (information $<$ 4) Then (TellDispensable)}
	\item{OsawaDiscard}
	\item{DiscardOldestFirst}
	\item{TellRandomly}
	\item{DiscardRandomly}
\end{itemize}

The first IfRule is designed as a hail Mary in the end game: if there is nothing left to lose, try to gain a point. This derives from human play, when typically during the end game we make random plays if we know there is a playable card somewhere in our hand. This rule is more accurate, as it uses all the information it has gathered to calculate probabilities. 

The second IfRule simply risks playing a card if there is a reasonable chance of its being safe.

The third IfRule is designed to try to provide more intelligent Tell conditions. If there is nothing useful to Tell and we are low on information, we set another agent up to be able to discard cards that are not needed. This means that the agents can burn through cards that are not helpful so as to try to obtain useful cards from the deck.

\subsubsection{Flawed}
\label{sec:agent-flawed}
This is an agent designed to be intelligent but with some flaws: it does not possess intelligent Tell rules, and has a risky Play rule as well. Understanding this agent is the key to playing well with it, because other agents can give it the information it needs to prevent it from playing poorly. The rules used in order are:
\begin{itemize}
	\item{PlaySafeCard}
	\item{PlayProbablySafeCard(0.25)}
	\item{TellRandomly}
	\item{OsawaDiscard}
	\item{DiscardOldestFirst}
	\item{DiscardRandomly}	
\end{itemize}

Giving information is the key to getting this agent to work intelligently. Without information, the intelligent rules can't fire, thereby leaving this agent to Tell randomly and Discard randomly --- not a great strategy. 

\subsubsection{Van den Bergh Rule}
\label{sec:agent-van}
This is the best rule-based agent from~\cite{vanaspects}. It was created by observing from human play that there are four main tasks:
\begin{itemize}
	\item{If I'm certain enough that a card is playable, Play it.}
	\item{If I'm certain enough that a card is useless, Discard it.}
	\item{Give a hint if possible.}
	\item{Discard a card.}
\end{itemize}
Van den Bergh {\em et al} used a \gls{ga} to evolve the best options for each section, resulting in the following rules as an implementation:
\begin{itemize}
	\item{IfRule (lives $>$ 1) Then (PlayProbablySafeCard(.6)) Else (PlaySafeCard)}
	\item{DiscardProbablyUselessCard(1.0)}
	\item{TellAnyoneAboutUsefulCard}
	\item{TellAnyoneAboutUselessCard}
	\item{TellMostInformation}
	\item{DiscardProbablyUselessCard(0.0)}
\end{itemize}

\subsubsection{\acrshort{mcs}}
\label{sec:agent-mcs}
This agent is a simple \gls{mcs} that uses a provided agent for the rollout phase. \gls{mcs} is a technique that uses the \gls{ucb} equation to select actions in a single step lookahead, with policy informed rollouts to evaluate those positions. It is essentially \gls{mcts} with a tree depth limit of one turn. In this paper, we name the agent \nameref{sec:agent-mcs}-[agent] to indicate which agent provided the rollout policy. For example, a \gls{mcs} agent using \nameref{sec:agent-iggi} as a policy would be named \nameref{sec:agent-mcs}-\nameref{sec:agent-iggi}. The agent has a one-second time limit to return a move.
\subsubsection{\acrshort{ismcts}}
\label{sec:agent-ismcts}
This agent uses a \gls{mcts} technique for handling games with partial observability as described in the paper by Cowling {\em et al}~\cite{Cowling2012}.

\gls{ismcts} is a modification to \gls{mcts} in which, on each iteration through the tree, the partially-observable game state is determinised into a possible fully-observable state. This state remains consistent for the selection, expansion, rollout and backpropagation phases before being replaced by a new determinisation. The implementation uses a time limit for returning moves of one second per move.

\subsubsection{Predictor \acrshort{ismcts}}
\label{sec:agent-predictor}
This agent was provided with a copy of each of the agents that it was paired with to use in its prediction. The predicted agents were initialised with random seeds: this corresponds to the predictor's having knowledge of each agent's overall strategy but no knowledge of its internal workings.

The Predictor \gls{ismcts} agent modifies the selection, expansion and rollout phases of \gls{mcts} when considering nodes for other agents turns. The modifications remove \gls{uct} for other agents' turns and replaces it with a query to the agent model to discover what that agent would do in that situation. The rollout phase is similarly modified. When making moves for its own turn, the predictor agent defaults to the legal random selection method used by \nameref{sec:agent-ismcts}. The implementation maintains the one-second-per-move limit of \nameref{sec:agent-ismcts}.

\section{Method}
\label{sec:method}

\subsection{Validation}
\label{sec:method:validation}
In order first to validate our framework and AI implementations, we performed experiments using reimplimentations of the Osawa and Van den Bergh agents. This involved recreating the experiments that they described in their papers and checking that we obtained similar results.

\subsection{Full Test}
\label{sec:method:full}
The set of agents under test contained a mix of current research on {\em Hanabi} as well as some rule-based agents of our own. There is also a mix of strong and poor agents for balance. We tested all the agents from this list: 
\begin{itemize}
	\item \nameref{sec:agent-legal_random}
	\item \nameref{sec:agent-outer}
	\item \nameref{sec:agent-iggi}		
	\item \nameref{sec:agent-piers}		
	\item \nameref{sec:agent-flawed}
	\item \nameref{sec:agent-van}		
	\item \nameref{sec:agent-mcs}-\nameref{sec:agent-legal_random}
	\item \nameref{sec:agent-mcs}-\nameref{sec:agent-iggi}
	\item \nameref{sec:agent-mcs}-\nameref{sec:agent-flawed}		
	\item \nameref{sec:agent-ismcts}
	\item \nameref{sec:agent-predictor}
\end{itemize}

In each experiment, one of the agents was selected from the list above and the remaining agents were selected as a group from the list below. For example, in the first experiment the \nameref{sec:agent-legal_random} agent would be alone among four \nameref{sec:agent-iggi} agents --- a concept we call {\em pairing}. The agents above were all paired in turn with:
\begin{itemize}
	\item \nameref{sec:agent-iggi}
	\item \nameref{sec:agent-internal}
	\item \nameref{sec:agent-outer}
	\item \nameref{sec:agent-legal_random}
	\item \nameref{sec:agent-van}
	\item \nameref{sec:agent-flawed}
	\item \nameref{sec:agent-piers}
\end{itemize}

200 random seeds were chosen, and for each seed every agent under test played two games with every agent with which it was paired. It did this for standard {\em Hanabi} rules with 2, 3, 4 and 5 players. Each agent under test played from a randomised position (first, second, third, fourth or fifth) determined by the seed. This ensured that each agent under test was in the same position for the same seed. Every agent therefore played $200(nSeeds) * 4(2, 3, 4 or 5 Players) * 7(nAgentPaired) * 2(reruns) = 11200$ games.

The configuration, final score and other basic state information were logged to a file upon completion of the game. The results were collated per agent and the mean score and number of turns taken were calculated. We also stored additional information about the final state of each game including the number of lives remaining and the information tokens remaining. When there are no lives remaining at the end of the game, this indicates that the players ran out of life tokens.

The full (human readable) game traces for each game are also stored, for evaluating agent behaviour and the effectiveness of strategies.

Finally, the configuration and results of each game are processed to obtain the mean score, mean number of moves per game and the mean remaining life and information tokens.

\section{Results}
\label{sec:results}

\subsection{\nameref{sec:method:validation}}
\renewcommand{\arraystretch}{1.4}
\begin{table*}[!htp]
	\centering
	\begin{tabularx}{12cm}{>{\bfseries}X | r | r | r | r}
		Agent & \textbf{Our Average} & \textbf{Their Average} & \textbf{N Games} & \textbf{N Players} \\ \hline
		\nameref{sec:agent-internal} & 10.12 (SD 1.98) & 10.97 (SD 1.94) & $10^2$ & 2\\
		\nameref{sec:agent-outer} & 13.83 (SD 2.23) & 14.53 (SD 2.24) & $10^2$ & 2\\
		\nameref{sec:agent-van} & 16.95 & 15.4 & $10^4$ & 3\\
	\end{tabularx}
	\caption{Table of results of validation tests}
	\label{tab:validate}
\end{table*}

The validation results are in \Cref{tab:validate}. The two Osawa agents obtained similar results in our system to those reported in the original paper. The \nameref{sec:agent-van} agent performed differently, appearing to be somewhat improved in our system.

\subsection{\nameref{sec:method:full}}
\begin{table}[!htp]
\centering
\begin{tabularx}{8cm}{>{\bfseries}X|r|r}
Agent & \textbf{Score} (2.d.p) & \textbf{Sem} (2.d.p) \\ \hline
\nameref{sec:agent-piers} & 11.18 & 0.06 \\
\nameref{sec:agent-mcs}-\nameref{sec:agent-iggi} & 10.97 & 0.06 \\
\nameref{sec:agent-iggi} & 10.96 & 0.06 \\
\nameref{sec:agent-van} & 10.88 & 0.06 \\
\nameref{sec:agent-predictor} & 10.74 & 0.06 \\
\nameref{sec:agent-outer} & 10.2 & 0.05 \\
\nameref{sec:agent-ismcts} & 5.9 & 0.04 \\
\nameref{sec:agent-mcs}-\nameref{sec:agent-legal_random} & 5.45 & 0.04 \\
\nameref{sec:agent-mcs}-\nameref{sec:agent-flawed} & 5.06 & 0.04 \\
\nameref{sec:agent-flawed} & 5.02 & 0.04 \\
\nameref{sec:agent-legal_random} & 4.59 & 0.04 \\
\end{tabularx}
\caption{Table of results with Score, Standard Error of the Mean and Ticks for each agent. Agents are sorted by score. N=11200}
\label{tab:results}
\end{table}

\begin{table}[!htp]
	\centering
	\begin{tabularx}{8cm}{>{\bfseries}X|r|r}
		Agent & \textbf{Score} (2.d.p) & \textbf{Sem} (2.d.p) \\ \hline
		\nameref{sec:agent-predictor} & 4.82 & 0.06 \\
		\nameref{sec:agent-iggi} & 3.26 & 0.06 \\
		\nameref{sec:agent-piers} & 3.24 & 0.06 \\
		\nameref{sec:agent-van} & 3.23 & 0.06 \\
		\nameref{sec:agent-mcs}-\nameref{sec:agent-iggi} & 3.21 & 0.06 \\
		\nameref{sec:agent-outer} & 2.96 & 0.05 \\
		\nameref{sec:agent-ismcts} & 1.8 & 0.04 \\
		\nameref{sec:agent-mcs}-\nameref{sec:agent-legal_random} & 1.78 & 0.04 \\
		\nameref{sec:agent-mcs}-\nameref{sec:agent-flawed} & 1.67 & 0.04 \\
		\nameref{sec:agent-legal_random} & 1.65 & 0.04 \\
		\nameref{sec:agent-flawed} & 1.59 & 0.04 \\
		
	\end{tabularx}
	\caption{Table of results with Score, Standard Error of the Mean and Ticks for each agent paired with Flawed. Agents are sorted by score. N=1600}
	\label{tab:results-flawed}
\end{table}

\Cref{tab:results} shows the full results for this test. Predictor \gls{ismcts} outperformed \gls{ismcts} in this experiment, with an average score of 10.74 versus \gls{ismcts}'s score of 5.9. \nameref{sec:agent-mcs} typically performed very similarly to the agent it was provided with for its rollouts; little benefit was apparent from using \gls{mcs} with these agents over simply using their rules in the first place. Overall, \nameref{sec:agent-piers} performed the best by a slim margin over \nameref{sec:agent-mcs}-\nameref{sec:agent-iggi}, \nameref{sec:agent-iggi} and \nameref{sec:agent-van}. The \nameref{sec:agent-flawed} agent was only a little better than \nameref{sec:agent-legal_random}. 

\begin{table}
\begin{tabularx}{\linewidth}{>{\bfseries}X | r | r | r| r}
	Agent & \textbf{2} & \textbf{3} & \textbf{4} & \textbf{5} \\ \hline
\nameref{sec:agent-flawed} & 3.52 & 4.69 & 5.43 & 6.45  \\
\nameref{sec:agent-iggi} & 11.76 & 11.29 & 10.71 & 10.09  \\
\nameref{sec:agent-ismcts} & 4.8 & 5.44 & 6.24 & 7.14  \\
\nameref{sec:agent-legal_random} & 1.68 & 4.3 & 5.83 & 6.53  \\
\nameref{sec:agent-mcs}-\nameref{sec:agent-flawed} & 3.61 & 4.72 & 5.43 & 6.48  \\
\nameref{sec:agent-mcs}-\nameref{sec:agent-iggi} & 11.79 & 11.34 & 10.68 & 10.09  \\
\nameref{sec:agent-mcs}-\nameref{sec:agent-legal_random} & 3.84 & 5.14 & 5.87 & 6.95  \\
\nameref{sec:agent-outer} & 10.55 & 10.64 & 9.99 & 9.62  \\
\nameref{sec:agent-piers} & \textbf{11.91} & 11.67 & 10.89 & 10.26  \\
\nameref{sec:agent-predictor} & 8.36 & \textbf{12.14} & \textbf{11.43} & \textbf{11.02}  \\
\nameref{sec:agent-van} & 10.55 & 11.76 & 10.91 & 10.29  \\
\end{tabularx}
\caption{Average scores for each agent over 2, 3, 4 and 5 player games sorted alphabetically. Bold scores indicate the highest score for that column.}
\label{tab:results-2player}
\end{table}

\section{Discussion}
\label{sec:discussion}
\begin{figure}[!h]
	\begin{subfigure}[b]{0.4\textwidth}
		\resizebox{\linewidth}{!}{
			\begin{tikzpicture}[->,>=stealth',level/.style={sibling distance = 5cm/#1,
				level distance = 1.5cm},scale=0.5,every node/.style={transform shape}] 
			\node [arn_n] {0}
			child{ 
				node [arn_r] (c1) {0}
				child{ node [arn_n] (c) {0} edge from parent node[above left] {Tell 1}}
				child{ node [arn_n] (d) {1} edge from parent node[above right] {Play 1}}     
				edge from parent node[above left] {Tell 1}                     
			}
			child{ node [arn_r] {0}
				child{ node [arn_n] (a) {0} edge from parent node[above left] {Tell 1}}
				child{ node [arn_n] (b) {1} edge from parent node [above right] {Play 1}
				}
				edge from parent node [above right]{Discard 1}
			}
			;
			\path (a) -- (b) node [midway] {$\cdots$};
			\path (c) -- (d) node [midway] {$\cdots$};
			\end{tikzpicture}}
		\caption{\nameref{sec:agent-ismcts}}
	\end{subfigure}\qquad\qquad
	\begin{subfigure}[b]{0.4\textwidth}
		\resizebox{\linewidth}{!}{
			\begin{tikzpicture}[->,>=stealth',level/.style={sibling distance = 5cm/#1,
				level distance = 1.5cm},scale=0.5,every node/.style={transform shape}] 
			\node [arn_n] {0}
			child{ 
				node [arn_r] (c1) {0}
				child{ node [arn_n] (c) {1} edge from parent node[left,midway] {Play 1}}
				edge from parent node[above left] {Tell 1}                     
			}
			child{ node [arn_r] {0}
				child{ node [arn_n] (a) {0} edge from parent node[above left] {Tell 1}}
				child{ node [arn_n] (b) {0} edge from parent node [above right] {Tell 2}
				}
				edge from parent node [above right]{Discard 1}
			}
			;
			\path (a) -- (b) node [midway] {$\cdots$};
			\end{tikzpicture}
		}
		\caption{\nameref{sec:agent-predictor}}
	\end{subfigure}
	\caption{Game trees from same state for both agents paired with \nameref{sec:agent-cautious} illustrating the difference in tree size between \nameref{sec:agent-ismcts} and \nameref{sec:agent-predictor} }
	\label{fig:mcts-tree}
\end{figure}
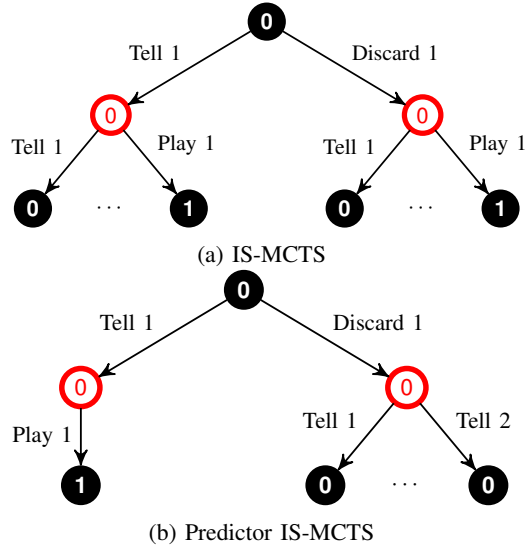

The Predictor \gls{ismcts} agent outperformed the \gls{ismcts} agent. This is mostly due to its better being able to take advantage of the effect of communication actions. As agents cannot see their own hands, the only way they gain information about their hands is via Tell actions; this then informs their decision process. When \gls{ismcts} appraises the moves of other agents in its tree, it considers all possible outcomes from that state. Some of these states will never occur in the real game because the paired agent would never select that action. The model that is available to Predictor \gls{ismcts} prunes the search to branches that are likely to occur in the game, resulting in more accurate statistics for the same number of iterations~(\Cref{fig:mcts-tree}). The more deterministic the model, the lower the branching factor for the tree will be. Smaller branching factors concentrate the rollouts, resulting in potentially more accurate statistics regarding those positions. More accurate statistics should result in more intelligent game play.

The Predictor \gls{mcts} really shows its benefit with the \nameref{sec:agent-flawed} agent as its partner. \Cref{tab:results-flawed} shows each agent when paired with \nameref{sec:agent-flawed}, with Predictor \gls{ismcts} in the clear lead ahead of other agents.

Interestingly, Predictor \gls{mcts}'s poor overall score appears to come largely from two-player games, for which it scores significantly lower than usual. This can be explained by the decreased rollout length present in these games. The more players in the game, the fewer random moves will be made in the rollouts (selecting random moves tends to end games very quickly with low scores, as exemplified by \nameref{sec:agent-legal_random}).

\Cref{tab:results-2player} shows all the agents' average scores over each player count. Most agents tend to follow one of two trends: either performing better when there are more players in the game, or performing worse. Those that improve are typically poor players, with each new player added to the game on average being better than them. Those that decline are the opposite: more players added means more poorer players in the team. Predictor \gls{ismcts} isn't the only agent to exhibit trouble with two player games, with \nameref{sec:agent-outer} experiencing some difficulty (despite having been designed for two-player games) and \nameref{sec:agent-van} displaying a more prominent drop in performance. In 3, 4 and 5 player games, the Predictor \gls{ismcts} is the best player from the set of agents.

\section{Conclusion}
\label{sec:conclusion}
In conclusion, we found that agent modelling improves playing strength for tree search algorithms such as \gls{mcts} in the game of {\em Hanabi}. These results are consistent with the findings of~\cite{barrett2011empirical}.

\section{Future work}
\label{sec:future}
There is a lot of scope for future work in this area. {\em Hanabi} has some additional variants in its rules that focus on the addition of a multi-coloured suit of cards. This suit also contains 3 1's, 2 2's, 3's and 4's as well as a single 5. The different variants are:

\begin{center}
\begin{tabularx}{\linewidth}{>{\bfseries}l X}
	Variant 1 &  Add the multi-coloured suit as a sixth suit to the game. Maximum score is boosted to 30. \\
	Variant 2 & Same as Variant 1, but only a single tile of each number from the multi-coloured suit is added to the game. \\
	Variant 3 & The multi-coloured suit now functions as a wild card in Tell actions, and cannot be directly called out. For example, if Player 1 tells Player 2 \{(M, 2), (Y, 2), (B, 5), (B, 3)\} about all the blues, then cards 1, 3 and 4 will be indicated. With this setup, the multi-coloured cards can only be identified by contradicting information given, requiring 3 pieces of information to fully identify one.\\ 
\end{tabularx}
\end{center}

Variant 1 would be simple to implement and test, but was omitted from this paper as being too off-topic. Variant 2 adds a little extra strategy, but is very similar to Variant 1. Variant 3 would require some additional work to implement, as well as appropriate modifications to the AI agents. 

The \nameref{sec:agent-predictor} has a number of limitations that we aim to address. The agent requires access to an accurate model of the co-operators in advance. It would be better if the agent could instead attempt to learn agent strategies based on observations in the game state. This would lead naturally to a more complicated agent that started with a more generic capability but was able to build models of its team members and update those models as games go on. Testing how much information is needed to learn enough to significantly improve the scores that a team achieve would then need to be done.

\section{Acknowledgements}
This work was funded by the EPSRC Centre for Doctoral Training in Intelligent Games \& Game Intelligence (IGGI) [EP/L015846/1]

The authors would like to thank members of the IGGI CDT for their assistance and support playing regular games of {\em Hanabi} with us.

\bibliographystyle{IEEEtran}
\bibliography{ms}

\end{document}